\documentclass[runningheads]{llncs}
\usepackage{graphicx}

\usepackage{graphicx}
\usepackage{amsmath}
\usepackage{amssymb}
\usepackage{booktabs}
\usepackage{multirow}
\usepackage{amsmath}

\usepackage[colorlinks=true,linkcolor=blue,urlcolor=blue,citecolor=blue]{hyperref}

\begin{document}

\raggedbottom

%
\title{Prompt Decoupling for Text-to-Image Person Re-identification}
%
%
\author{Weihao li\inst{1} \and
Lei Tan\inst{1} \and
Pingyang Dai\inst{2} \and
Yan Zhang\inst{3}}
%
%
%
\maketitle              

\begin{abstract}

Text-to-image person re-identification (TIReID) aims to retrieve the target person from an image gallery via a textual description query. Recently, pre-trained vision-language models like CLIP have attracted significant attention and have been widely utilized for this task due to their robust capacity for semantic concept learning and rich multi-modal knowledge. However, recent CLIP-based TIReID methods commonly rely on direct fine-tuning of the entire network to adapt the CLIP model for the TIReID task. Although these methods show competitive performance on this topic, they are suboptimal as they necessitate simultaneous domain adaptation and task adaptation. To address this issue, we attempt to decouple these two processes during the training stage. Specifically, we introduce the prompt tuning strategy to enable domain adaptation and propose a two-stage training approach to disentangle domain adaptation from task adaptation. In the first stage, we freeze the two encoders from CLIP and solely focus on optimizing the prompts to alleviate domain gap between the original training data of CLIP and downstream tasks. In the second stage, we maintain the fixed prompts and fine-tune the CLIP model to prioritize capturing fine-grained information, which is more suitable for TIReID task. Finally, we evaluate the effectiveness of our method on three widely used datasets. Compared to the directly fine-tuned approach, our method achieves significant improvements.

\end{abstract}
\section{Introduction}

Text-to-image person re-identification (TIReID) aims to retrieve images of a target person from an image gallery using natural language descriptions as queries. Traditional image person re-identification~\cite{tradition} assumes that an image of the target person can be obtained as the query during the retrieval processing. However, the query image is not always available in some cases. In comparison to images, text descriptions of a person are easier to obtain and fulfill practical application requirements. Consequently, TIReID has gained significant attention in recent years and has found extensive applications in smart cities and security surveillance.


TIReID is a fine-grained matching task due to the high similarity in human body structures, requiring identification through fine-grained attributes. Consequently, the major challenge of this task lies in mining and aligning fine-grained information across modalities. Previous studies employed single-modality pre-trained models, such as ResNet~\cite{resnet} and Bert~\cite{bert}, to initialize their backbones. Subsequently, the features were projected for better alignment. Existing methods force the feature representations of different modalities with the same semantics to align in the shared feature space, which can lead to intra-modal information distortion~\cite{dist1,dist2}.

Recently, large-scale Vision-Language Pre-training (VLP) models have provided new solutions for the TIReID task. VLP models are trained on large-scale datasets containing text-image pairs and possess robust feature joint representation capabilities. Han et al.~\cite{han} applied momentum contrastive learning to CLIP with the goal of learning better latent space feature representations (with minimal intra-class distance and maximal inter-class distance) on smaller TIReID datasets. CLIP-Driven~\cite{cfine} proposed a multi-modal interaction module based on CLIP to mine fine-grained information of image-text pairs. It is worth noting that when applying VLP models to downstream tasks, in addition to task adaptation, domain adaptation is also required. Specifically, there is a certain gap between the downstream data domain and the original training data domain of VLP, such as completely contrasting textual description styles as well as different scenes depicted in the images across the two domains. However, these studies attempt to directly apply the valuable knowledge of CLIP to the subsequent TIReID task, necessitating simultaneous task adaptation and domain adaptation, which can potentially lead to a loss of valuable knowledge.

To tackle the aforementioned issues, we employ prompts to bring the downstream data domain closer to the original training data domain of the CLIP. Additionally, we adopt a two-stage training strategy to separate domain adaptation from task adaptation. Specifically, during the first stage, we only optimize the prompts by contrast loss, thereby facilitating domain adaptation. In the second stage, these well-learned prompts are frozen and fine-tune the encoders, enabling the model to focus on fine-grained information for task adaptation. In contrast to previous methods, which tackles domain and task adaptation simultaneously, our method facilitates a more effective transfer of the powerful knowledge of VLP to the downstream TIReID task. We summarize our primary contributions as follows:

\begin{enumerate}
  \item We propose a prompt-tuning strategy for domain adaptation. To our knowledge, we are the first to introduce prompts to bridge the domain gap between the downstream task data and the training data of CLIP.
  \item We propose a two-stage training strategy that separates domain adaptation and task adaptation, which facilitating the application of CLIP's powerful knowledge to downstream TIReID task.
  \item We conducted extensive experiments on three popular TIReID datasets to validate the efficacy of our method. In comparison to directly full fine-tuning, which necessitates simultaneous domain and task adaptation, our method exceeds +3.4\%, +4.19\% and +2.6\% on Rank-1.
\end{enumerate}



\section{Related Work}
\subsection{Text-to-image person Re-identification}

This is a cross-modal retrieval task and was initially proposed by Li et al.~\cite{cuhk}. 
The primary challenge of TIReID task lies in capturing fine-grained features and aligning cross-modal features in a latent space. Early approaches primarily employed a unimodal encoder to separately extract features from text and image. Subsequently, these features were projected onto a shared latent space using a projection layer to achieve global or local feature alignment.~\cite{dualpath,timam} aligns global features at the expense of sacrificing numerous fine-grained features, which is suboptimal for this fine-grained cross-modal retrieval task. Conversely,~\cite{cmpc,vitaa,rstp,ssan,isanet} concentrate on align local features to capture more fine-grained information, resulting in improved performance.~\cite{asymmetric} discovered that there exists an asymmetric correspondence between word and global person feature. However, these method extracts features through unimodal feature extractor and then projects features onto a joint latent space can easily lead to training confusion~\cite{dist1}. Recently, the emergence of various VLP models has introduced new solutions for the TIReID task. The objective of pre-training visual-language models is to acquire a multimodal model with robust representation capabilities by leveraging a large-scale dataset of text-image pairs. This pre-trained model can subsequently be fine-tuned or directly employed for downstream tasks. CLIP~\cite{clip} is a notable example. Han et al.~\cite{han} were the pioneers in applying CLIP to the TIReID task. They introduced a cross-modal momentum contrastive learning scheme that enables the model to learn more distinctive features on relatively small TIReID datasets. CFine~\cite{cfine} proposed a sophisticated framework to mine fine-grained information from text and images, thereby facilitating more accurate retrieval.

\subsection{Prompt Learning}

This technique was initially used in NLP~\cite{firstprompt} to incorporate prompt structure into the input of pre-trained language model so that the model can better understand downstream task. Although manually defined prompts~\cite{mp1} have improved the performance on downstream tasks, they require a significant amount of expert knowledge and may not adapt well to the language model. To tackle this issue, certain researchers have designed prompts as learnable continuous vectors that can be directly optimized through gradient backpropagation based on specific tasks. ViT~\cite{vit} has been directly applied Transformer-based architecture as visual encoder in image classification tasks, so researchers have also attempted to use prompts in visual tasks. VPT~\cite{vpt} employed prompt learning to better adapt to visual downstream tasks instead of full fine-tune the pre-trained visual model, with even got better performance. CoOp~\cite{coop} has further incorporated the prompt mechanism into VLP models by replacing fixed context (such as ``a photo of") with a learnable prompt for image recognition tasks, achieving good performance. Additionally, some works~\cite{domain1,domain2} have explored the utilization of prompts as domain adaptor.

\section{Methods}

\begin{figure}
\vspace{-1em}
\includegraphics[width=\textwidth]{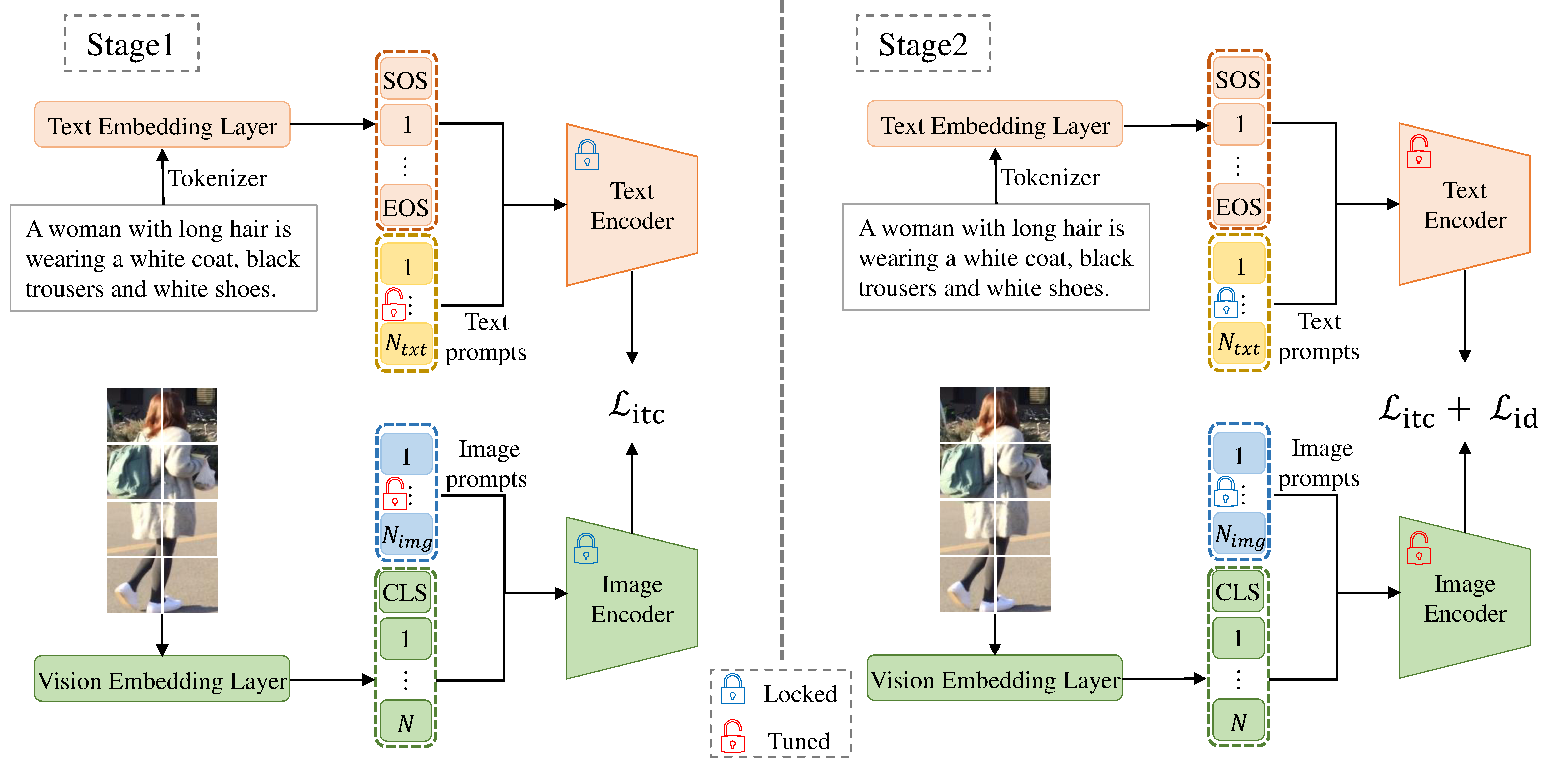}
\caption{Overview of our approach, which adopts a two-stage training strategy. In the first stage (left), we optimize the prompts for domain adaptation while keeping CLIP frozen. In the second stage (right), we freeze the prompts and fine-tune CLIP for task adaptation.} \label{fig:model}
\vspace{-3em}
\end{figure}

In this paper, we propose a prompt-based method for domain adaptation when applying CLIP to downstream TIReID task, and correspondingly adopt a two-stage training strategy to separate domain adaptation and task adaptation. Fig.\ref{fig:model} illustrates the overall model framework, we will elaborate on the design of the prompt in Sect.\ref{sec:3.1}, and introduce the two-stage training strategy as well as the corresponding objective function in Sect.\ref{sec:3.2}.

\subsection{Prompt for domain adaptation}
\label{sec:3.1}

Firstly, we provide a brief overview of CLIP-ViT-B/16, which serves as our backbone. This architecture comprises a text encoder, denoted as $\mathcal{T}(\cdot)$, and an image encoder, denoted as $\mathcal{I}(\cdot)$. Both encoders employ a 12-layer Transformer structure. In the text encoder, $\mathcal{T}(\cdot)$ initially converts each word in a text description into a unique numeric ID by lower-cased byte pair encoding (BPT) with a vocabulary size of 49152. To signify the start and end of the sequence, the tokens are then enclosed with [SOS] and [EOS] tokens, respectively. Subsequently, an embedding layer maps all tokens to the Transformer's embedding space, and positional embeddings are added to the tokenized text as input to the Transformer. We define the input of $($i+1$)$-th text Transformer block as $\boldsymbol T_i = \left\{{t}_{sos}^i, {t}_1^i, {t}_2^i, \ldots,{t}_{eos}^i\right\}$, the ${t}_{eos}^{12}$ after a linear project to serve as the global text feature. The image encoder follows a similar transformer structure as the text encoder, with the difference lying in the additional processing as patchify required for images. Given an image $I \in R^{H \times W \times C}$, where $(H,W)$ represents the image resolution, $C$ is the number of channels, the image is divided into a sequence of $N={HW}/{P^2}$ non-overlapping patches, where $P$ refers to the patch size. The patch sequence is then projected onto the embedding space, and an extra [CLS] token is appended. Similar to the text encoder, we define $\boldsymbol V_i = \left\{{v}_{cls}^i, {v}_1^i, {v}_2^i, \ldots, {v}_{N}^i\right\}$ as the input of $($i+1$)$-th vision Transformer block, and $v_{cls}^{12}$ is projected into the joint image-text embedding space, serving as the global image feature.

In this paper, we employ prompt learning for domain adaptation when introducing CLIP to downstream tasks. As shown in Fig.\ref{fig:model}, we concatenate prompts in the input sequence of two encoders to learn to bridge the domain gap between the downstream data and the original CLIP training data. Prompts are defined as continuous learnable vectors, that keep the same dimension as the embedding space of the Transformer block. It is worth mentioning that position embeddings are not added to prompts, rendering the insertion position inconsequential. Thus, we can adopt a simple concatenation method. More specifically, text prompts and image prompts are denoted as $\boldsymbol P_{txt} = \left\{{p}_1^{txt}, {p}_2^{txt}, \ldots, {p}_{N_{txt}}^{txt}\right\}$, $\boldsymbol P_{img} = \{{p}_1^{img}, {p}_2^{img}, \ldots, {p}_{N_{img}}^{img}\}$, respectively. The superscript of each prompt shows which encoder it is added to. $N_{img}\left(N_{txt}\right)$ denotes the length of the image(text) prompt. $\boldsymbol P_{txt}$ is directly appended after $\boldsymbol{t}_{eos}^0$, 
so $\boldsymbol T_0$ changes as:
\begin{equation}
\boldsymbol T_0 = \{{t}_{sos}^0, {t}_1^0, {t}_2^0, \ldots,{t}_{eos}^0,{p}_1^{txt}, {p}_2^{txt}, \ldots, {p}_{N_{txt}}^{txt} \}.
\end{equation}

Here, the superscript of $t$ means the layer that prompts located. As to the image encoder, The $\boldsymbol P_{img}$ is inserted before the [cls] token, so the input of the image encoder denotes as: 

\begin{equation}
\boldsymbol V_0 = \{{p}_1^{img}, {p}_2^{img}, \ldots, {p}_{N_{img}}^{img}, {v}_{cls}^0, {v}_1^0, {v}_2^0, \ldots, {v}_{N}^{0}\}.
\end{equation}

It is worth mentioning that we also tried prompt deep~\cite{vpt}, which introduces prompts additional input spaces within the Transformer blocks apart from $\boldsymbol T_0, \boldsymbol V_0$, the performance is not better.


\subsection{Two-stage training strategy}
\label{sec:3.2}
Following the design of the prompt, we further introduce a two-stage training paradigm that separates domain adaptation and task adaptation, which enhances the applicability of CLIP to downstream TIReID task.

\subsubsection{The first training stage.} We lock the entire CLIP at this stage and optimize the learnable prompts, as shown in the left half of Fig.\ref{fig:model}, we optimize prompts through the contrastive loss same as CLIP. Specifically, we project ${t}_{eos}^{12}$ to obtain the text feature ${f}_{y}^{txt}$, and project ${v}_{cls}^{12}$ to obtain the final image feature ${f}_{y}^{img}$. The subscript represents the person ID, so the text-to-image contrastive loss can be expressed as follows:

\begin{equation}
\mathcal{L}_{t 2 i}\left(y_i\right)=\frac{-1}{\left|P\left(y_i\right)\right|} \sum_{p \in P\left(y_i\right)} \log \frac{\exp \Big(D\big(f_p^{txt}, f_{y_i}^{img}\big)\Big)}{\sum_{a=1}^B \exp \Big(D\big(f_a^{txt}, f_{y_i}^{img}\big)\Big)}
\end{equation}

$D(\cdot)$ refers to the similarity function that evaluates the distance between text and image. In this work, we use cosine similarity. $P{(y_i)}$ indicates the positive set of $T_{y_i}$ in a mini-batch, $|\cdot|$ is its cardinality, image to text contrastive loss can be denoted in the same way. So the contrastive loss between text and image is given as: 

\begin{equation}
\mathcal{L}_{itc}=\mathcal{L}_{t2i}+\mathcal{L}_{i2t}    
\label{eq:itc}
\end{equation}



The gradients are back-propagated through the $\mathcal{I}(\cdot)$ and $\mathcal{T}(\cdot)$ to optimize the prompts, by this way, we reduce the gap between the downstream data domain and the original CLIP training data domain. Therefore, during the subsequent fine-tuning stage, the CLIP can focus exclusively on task adaptation.


\subsubsection{The second training stage.} 

In the first stage, domain adaptation is completed. During this stage, we keep the well-learned prompts fixed and solely fine-tune the two encoders for downstream task adaptation. Specifically, We fine-tune CLIP to pay more attention to fine-grained information, leveraging the inherent capability of the Transformer architecture to increase the weight of fine-grained information during self-attention. As shown in the right half of Fig.\ref{fig:model}, we concatenate the well-learned prompts with the embeddings of descriptions/images and input them into Transformer blocks. We also use Eq.\ref{eq:itc} to learn a better feature representation. In addition, to increase intra-class compactness, we add a cross-modal ID loss, which is represented by the following formula:

\begin{equation}
\mathcal{L}_{i d}=\sum_{k=1}^N-q_k \log \left(p_k\right)    
\end{equation}

Here, $p$ represents the logits distribution obtained by the classifier, $q$ represents the ground truth distribution (often a one-hot vector), and $N$ is the total number of instances. This loss function deviates from the conventional classification loss as both text and image pass through the same classifier to obtain their respective logits. The total loss in the second stage is given by the following equation. It is important to highlight that id loss is a strong constraint,
so we set the hyperparameter $\lambda$ to 0.1.

\begin{equation}
\mathcal{L}_{total}=\mathcal{L}_{itc}+\lambda\mathcal{L}_{id}    
\end{equation}

The utilization of the aforementioned two-stage training strategy enables the effective separation of domain adaptation and task adaptation, thereby facilitating the application of CLIP's powerful knowledge to downstream tasks.


\section{Experiments}
\subsection{Datasets and Evaluation Protocol}
We evaluate our method in three popular TIReID datasets, including CUHK-PEDES~\cite{cuhk}, ICFG-PEDES~\cite{ssan}, and RSTPReid~\cite{rstp}. It is important to note that CUHK-PEDES and RESPReid have two descriptions per image on average, while ICFG-PEDES has a single textual description per image. The details of these datasets are summarized in Table~\ref{tab:dataset}. For evaluation metrics, we employ the most widely used Cumulative Matching Characteristic(CMC) and mean Average Precision(\emph{m}AP) as evaluation metrics. In addition, we also adopt the Inverse Negative Penalty(\emph{m}INP)~\cite{minp} as an additional retrieval criterion.


\begin{table}[htbp]
  \caption{Details of three popular datasets.}
  \centering
  \resizebox{0.9\textwidth}{!}{
  \begin{tabular}{l|ccc|ccc|ccc}
  \toprule
  \multirow{2}{*}{Subset}  &\multicolumn{3}{c|}{CUHK-PEDES} &\multicolumn{3}{c|}{ICFG-PEDES} &\multicolumn{3}{c}{RSTPReid} \\ \cline{2-10} 
                  &ID  &Image  &Caption     &ID  &Image  &Caption   &ID  &Image  &Caption\\ \hline
  Training set    &11003  &34054  &68126  &3102   &34674   &34674   &3701  &18505  &37010  \\
  Validation set  &1000   &3058   &6158   &-       &-      &-       &200  &1000  &2000  \\
  Test set        &1000   &3074   &6156   &1000   &19848   &19848   &200  &1000  &2000  \\ 
  \bottomrule
  \end{tabular}}
  \label{tab:dataset}
  \vspace{-4mm}
\end{table}

\subsection{Implementation Details}
We conducted all experiments using PyTorch~\cite{pytorch} on a single RTX3090 GPU with CLIP-ViT-B/16 as the backbone model. For text encoder, the maximum length of the tokens is set to 77, text descriptions in the dataset are no longer than 50 words, the text Transformer block had a width of 512, and the number of heads is set to 8. For image encoder, the images in dataset are resized to $384\times128$, the patch size is $16\times16$, so there are 193 (add a [cls] token) patches, so we first to resized the vision position embedding from 197 to 193, the hidden size of vision Transformer layer is 768, we final projected the dimension of features from 768 to 512. Random horizontally flipping, random crop with padding, and random erasing are employed for image data augmentation. Both text encoder and image encoder contain 12 Transformer layers and all prompts are randomly initialized with xavier~\cite{xavier} uniform initialization scheme. During training, we set our base learning rate as $1 \times 10^{-5}$, and spent 5 warm-up epochs linearly increasing the learning rate from $1 \times 10^{-6}$ to $1 \times 10^{-5}$, for classifier, it was set as $5\times$ the base learning rate, Adam optimizer~\cite{adam} with cosine learning rate decay is used to train our model, the training configuration remained consistent across both training stages, we set each stage 60 epochs for full convergence.

\begin{table}[h]
  \caption{Performance comparisons on CUHK-PEDES. ``G" and ``L" stand for global-matching/local-matching method.}
  \centering  
  \resizebox{0.9\textwidth}{!}{
  \begin{tabular}{l|c|ccccc}
  \toprule
  Method                                &Type      &Rank-1            &Rank-5         &Rank-10        &\emph{m}AP         &\emph{m}INP                \\
  \hline

  Dual Path~\cite{dualpath}         &G              &44.40             &66.26          &75.07          &-              &-              \\
  CMPM/C~\cite{cmpc}            &L              &49.37             &-              &79.27          &-              &-              \\
  TIMAM~\cite{timam} &G              &54.51             &77.56          &79.27          &-              &-              \\
  ViTAA~\cite{vitaa}             &L              &54.92             &75.18          &82.90          &51.60          &-              \\
  DSSL~\cite{rstp}                &L              &59.98             &80.41          &87.56          &-              &-              \\
  SSAN~\cite{ssan}       &L              &61.37             &80.15          &86.73          &-              &-              \\
  ISANet~\cite{isanet}             &L              &63.92             &82.15          &87.69          &-              &-              \\
  LBUL~\cite{lbul}               &L              &64.04             &82.66          &87.22          &-              &-              \\
  Han et al.~\cite{han}          &G              &64.08             &81.73          &88.19          &60.08          &-              \\
  LGUR~\cite{lgur}           &L              &65.25             &83.12          &89.00          &-              &-              \\
  IVT~\cite{ivt}                  &G              &65.59             &83.11          &89.21          &-              &-              \\
  CFine~\cite{cfine}               &L              &\underline{69.57} &85.93          &91.15          &-              &-              \\
  \hline
  \textbf{Baseline (CLIP-ViT-B/16)}     &G              &68.20  &\underline{86.47} &\underline{91.47} &\underline{61.12} &\underline{44.86}\\
  \textbf{Ours}                  &G              &\textbf{71.59} &\textbf{87.95} &\textbf{92.45} &\textbf{65.03} &\textbf{49.97}    \\
  \bottomrule
  \end{tabular}}%
  \label{tab:cuhk}
  \end{table}

\subsection{Comparison with State-of-the-Art Methods}
In this section, we compared our method with the state-of-the-art methods on three widely used public benchmark datasets. Baseline is directly full fine-tune CLIP with InfoNCE loss~\cite{infonce}. 

\begin{table}[h]
  \caption{Performance comparisons on ICFG-PEDES.}
  \centering  
  \resizebox{0.9\columnwidth}{!}{
  \begin{tabular}{l|c|ccccc}
    \toprule
    Method                            &Type   &Rank-1       &Rank-5      &Rank-10       &\emph{m}AP        &\emph{m}INP         \\
    \hline
    Dual Path~\cite{dualpath}     &G      &38.99        &59.44       &68.41         &-          &-            \\
    CMPM/C~\cite{cmpc}        &L      &43.51        &65.44       &74.26         &-          &-            \\
    ViTAA~\cite{vitaa}         &L      &50.98        &68.79       &75.78         &-          &-            \\
    SSAN~\cite{ssan}   &L      &54.23        &72.63       &79.53         &-          &-            \\
    IVT~\cite{ivt}              &G      &56.04        &73.60       &80.22         &-          &-            \\
    ISANet~\cite{isanet}         &L      &57.73        &75.42       &81.72         &-          &-            \\
    CFine~\cite{cfine}           &L &\underline{60.83} &\underline{76.55} &\underline{82.42} &-  &-          \\

    \hline
    \textbf{Baseline (CLIP-ViT-B/16)} &G      &56.74        &75.72	     &82.26	        &31.84	    &5.03         \\
    \textbf{Ours}             &G &\textbf{60.93} &\textbf{77.96} &\textbf{84.11} &\textbf{36.44} & \textbf{7.79} \\
    \bottomrule
    
  \end{tabular}}
  \label{tab:icfg}
  \vspace{-4mm}
\end{table}

\subsubsection{Performance Comparisons on CUHK-PEDES.}
As presented in Table~\ref{tab:cuhk}, It is worth mentioning that the Baseline achieves 68.20\%, 61.12\% on Rank-1 and \emph{m}AP, respectively, has already achieved recent state-of-the-art method CFine~\cite{cfine}, our method outperforms +2.02\%, +2.02\%, +1.3\% than CFine on Rank-1, RanK-5 and Rank-10, respectively, reach 71.59\%, 87.95\%, 92.45\% on these metrics.

\begin{table}[h]
  \caption{Performance comparisons on RSTPReid.}
  \centering  
  \resizebox{0.9\columnwidth}{!}{
  \begin{tabular}{l|c|ccccc}
    \toprule
    Method                            &Type   &Rank-1       &Rank-5      &Rank-10       &\emph{m}AP        &\emph{m}INP         \\
    \hline
    DSSL~\cite{rstp}            &G      &39.05        &62.60       &73.95         &-          &-            \\
    SSAN~\cite{ssan}   &L      &43.50        &67.80       &77.15         &-          &-            \\
    LBUL~\cite{lbul}           &L      &45.55        &68.20       &77.85         &-          &-            \\
    IVT~\cite{ivt}              &G      &46.70        &70.00       &78.80         &-          &-            \\
    CFine~\cite{cfine}           &L      &50.55        &72.50       &81.60         &-          &-            \\
    \hline
    \textbf{Baseline (CLIP-ViT-B/16)} &G &\underline{54.05}	&\textbf{80.70} &\textbf{88.00} &\underline{43.41} &\underline{22.31}\\
    \textbf{Ours}             &G &\textbf{56.65} &\underline{77.40} &\underline{84.70} &\textbf{45.27} & \textbf{26.02} \\
    \bottomrule
  \end{tabular}}
  
  \label{tab:rstp}
  \vspace{-4mm}
\end{table}

\subsubsection{Performance Comparisons on ICFG-PEDES.}The results of our method on this dataset as shown in Table~\ref{tab:icfg}, it achieved an accuracy of 60.93\%, 77.96\%, and 84.11\% on Rank-1, Rank-5, and Rank-10, respectively, which is outperforms the Baseline with +4.19\%, +2.24\%, +1.85\%. We can observe that the \emph{m}INP metric, which calculates the total number of correct matches divided by the index of the last correct match, is quite low. This can be attributed to the fact that each image in this dataset has only one corresponding textual description, resulting in a relatively small number of correct matches.

\subsubsection{Performance Comparisons on RSTPReid.}
The results as shown in Table~\ref{tab:rstp}, we can see that Baseline exceed  +3.5\%, +8.2\%, +6.4\% than Cfine on Rank-1, Rank-5, Rank-10, respeatively, Compared with Baseline, our method further improved +2.6\%, +3.71\% on Rank-1 and \emph{m}INP.

In summary, the experimental results on these three challenging TIReID datasets demonstrate the effectiveness of our proposed method. 
The results of the baseline method highlight the generalization and robustness of VLP, which is increasingly dominating the field.

\subsection{Ablation Studies}
To comprehensively demonstrate the impact of different configurations in our method, we conducted extensive ablation studies on the CUHK-PEDES dataset. In this section, we first evaluated the contribution of prompt and the two-stage training strategy. Subsequently, we analyzed the effect of prompt length, which is the only additional hyperparameter in our model.

\begin{table}[h]
  \caption{Ablation study on prompt and two-stage on CUHK-PEDES.}
  \centering
  \resizebox{0.9\textwidth}{!}{
  \begin{tabular}{c|c|cc|ccccc}
  \toprule
  No.  &Prompt    &One-stage   &Two-stage   &Rank-1  &Rank-5  &Rank-10                  &\emph{m}AP      &\emph{m}INP \\ \hline
  0    &          &\checkmark  &            &68.20   &86.47   &91.47                    &61.12    &44.86\\
  1    &\checkmark&\checkmark  &            &70.66   &87.31   &92.21           &64.29    &49.22\\ 
  2   &\checkmark&     &\checkmark&\textbf{71.59}&\textbf{87.95}&\textbf{92.45}&\textbf{65.03}&\textbf{49.97}\\

  \bottomrule
  \end{tabular}
  }
  \label{tab:ps}
  \vspace{-10mm}
\end{table}

\subsubsection{Prompt and Two-Stage Training.} Our method adopts a two-stage training strategy aiming to separate domain adaptation and task adaptation and utilizing prompts to alleviate the domain gap between TIReID data and VLP training data. As shown in Table~\ref{tab:ps}, in No.1, prompts are added while employing a one-stage training strategy where the prompt and two encoders are optimized simultaneously. Compared with No.1 and No.0(Baseline), No.1 exhibited an improvement in Rank-1, \emph{m}AP, and \emph{m}INP accuracy by +2.46\%, +3.17\%, and +4.36\% on the CUHK-PEDES. These results clearly demonstrate that prompt learning can be beneficial for domain adaptation. The efficacy of two-stage training strategy is revealed via the experimental results of No.1 vs No.2, where both set prompt length as 2, two-stage training strategy outperforms one-stage training strategy by +0.93\%, +0.74\% on Rank-1 and \emph{m}AP.

\begin{figure}[h]
\centering
\vspace{-4mm}
\includegraphics[width=0.9\textwidth]{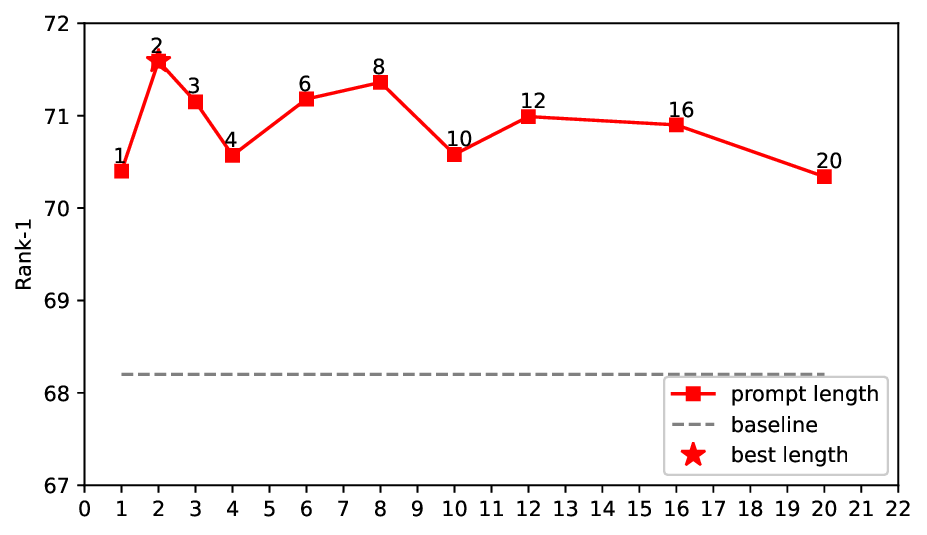}
\vspace{-4mm}
\caption{Ablation study on prompt length on CUHK-PEDES.} \label{fig:prompt_length}
\vspace{-6mm}
\end{figure}

\subsubsection{Prompt length.}We performed additional ablation experiment on the prompt length $N_p$, which was set to be the same for both text and image. As shown in Fig.\ref{fig:prompt_length}, the range of $N_p$ is $\left\{1, 2, 3, 4, 6, 8, 10, 12, 16, 20\right\}$, we fixed the prompt dropout rate of 0.3, setting the prompt length to 2 yielded good performance. Besides, we also investigated the effects of using different lengths for text prompts and image prompts, the results as shown in Table~\ref{tab:prompt_length}, we keep one length of prompt equal to 8 as pivot, adjusting the another length, the results make we get the same conclusion as VPT~\cite{vpt} that is the choice of prompt length is dependent on the downstream task and the number of classes in the dataset. The number of text instances and image instances may vary, but the classes is same, so prompts of same length for two encoders may be more appropriate.

\begin{table}[h]
  \caption{Ablation study on different length for text/image prompt on CUHK-PEDES.}
  \centering
  \resizebox{0.9\textwidth}{!}{
  \begin{tabular}{c|cc|ccccc}
  \toprule
  No.  &Text-prompts&Image-prompts &Rank-1     &Rank-5  &Rank-10        &\emph{m}AP      &\emph{m}INP \\ \hline
  0    &8            &8               &\textbf{71.36}   &\textbf{87.47} &91.98     &64.61    &49.61\\
  1    &8            &6               &70.86   &87.02   &91.85          &\textbf{64.63}    &\textbf{49.70}\\ 
  2    &8            &10              &70.42   &87.15   &92.09          &64.03    &48.98\\
  3    &6            &8               &71.13   &87.43   &\textbf{92.37} &64.32    &49.02\\
  4    &10           &8               &70.55   &87.17   &92.06 &64.52    &49.73\\
  \bottomrule
  \end{tabular}
  }
  \label{tab:prompt_length}
  \vspace{-4mm}
\end{table}

We primarily conducted ablation experiments focusing on the two main aspects discussed earlier. Interestingly, we observed that omitting the addition of position embeddings to the prompt led to improved performance. This may be that adding position encode to the prompt may disrupt the inherent powerful knowledge of the CLIP. Transformer has powerful global modeling capabilities if without position code, therefore placing the prompt anywhere has minimal impact on the results.

\section{Conclusion}
This paper proposed a two-stage training strategy that effectively separates domain adaptation and task adaptation when applying the CLIP for downstream TIReID task. Specifically, during the first stage, we optimize learnable prompts to align the downstream data domain with the original training data domain of CLIP, facilitating domain adaptation. In the second stage, we fine-tune CLIP to enhance its focus on fine-grained information, thereby facilitating task adaptation. This straightforward approach yields excellent performance on three widely used datasets, demonstrating its effectiveness. We hope that our work will inspire related research on transferring large pre-trained models to downstream tasks.

\bibliographystyle{splncs04}
\bibliography{ref}

\end{document}